\documentclass[11pt,letterpaper,logo,twocolumn]{style}

\usepackage[numbers]{natbib}
\usepackage{graphicx}
\usepackage{booktabs}
\usepackage{amsmath,amsfonts,amssymb}
\usepackage{subcaption}
\usepackage{multirow}
\usepackage{colortbl}
\usepackage{listings}
\usepackage{xparse}
\usepackage{fontawesome5}
\usepackage{threeparttable}

\graphicspath{{./}{fig/}{figures/}{plot/}{pdf/}{table/}}
\usepackage{amsthm}
\usepackage{tcolorbox}
\tcbuselibrary{skins,breakable}
\tcbuselibrary{listingsutf8}
\usepackage{titletoc}
\usepackage{pifont}
\usepackage{mathtools}
\usepackage{bbm}
\usepackage{makecell}
\usepackage{adjustbox}

\usepackage{algorithmic}
\usepackage{algorithm}
\usepackage{caption}
\usepackage{float}
\usepackage{svg}

\title{Reasoning in the Dark: Interleaved Vision-Text Reasoning in Latent Space}
\runningtitle{Reasoning in the Dark: Interleaved Vision-Text Reasoning in Latent Space}
\PublicDate{2026-1-29}

\author{%
  {\Authfont
    \textbf{Chao Chen}\textsuperscript{1} \quad
    \textbf{Zhixin Ma}\textsuperscript{2} \quad
    \textbf{Yongqi Li}\textsuperscript{1}\advisor \quad
    \textbf{Yupeng Hu}\textsuperscript{3} \quad
    \textbf{Yinwei Wei}\textsuperscript{3} \quad
    \textbf{Wenjie Li}\textsuperscript{1} \quad
    \textbf{Liqiang Nie}\textsuperscript{4}
  }\\
  {\Affilfont
\textsuperscript{1} \mbox{The Hong Kong Polytechnic University} 
\textsuperscript{2} \mbox{Singapore Management University} 
\textsuperscript{3} \mbox{Shandong University} \\
\textsuperscript{4} \mbox{Harbin Institute of Technology (Shenzhen)}

    \texttt{ochenchaoo@outlook.com, liyongqi0@gmail.com}
  }
}

\begin{document}

% ============================================================
% Abstract
% ============================================================
\begin{abstract}
Multimodal reasoning aims to enhance the capabilities of MLLMs by incorporating intermediate reasoning steps before reaching the final answer. It has evolved from text-only reasoning to the integration of visual information, enabling the thought process to be conveyed through both images and text. Despite its effectiveness, current multimodal reasoning methods depend on explicit reasoning steps that require labor-intensive vision-text annotations and inherently introduce significant inference latency. To address these issues, we introduce multimodal latent reasoning with the advantages of multimodal representation, reduced annotation, and inference efficiency. To facilitate it, we propose Interleaved Vision-Text Latent Reasoning (IVT-LR), which injects both visual and textual information in the reasoning process within the latent space. Specifically, IVT-LR represents each reasoning step by combining two implicit parts: latent text (the hidden states from the previous step) and latent vision (a set of selected image embeddings). We further introduce a progressive multi-stage training strategy to enable MLLMs to perform the above multimodal latent reasoning steps. Experiments on M$^3$CoT and ScienceQA demonstrate that our IVT-LR method achieves an average performance increase of 5.45\% in accuracy, while simultaneously achieving a speed increase of over 5 times compared to existing approaches.

\end{abstract}

% ============================================================
% Links (add more if needed)
% ============================================================
\newcommand{\TitleLinks}{%
\centering
    \vspace{6pt}
    {\noindent\absfont\fontsize{11}{13}\selectfont
    \faGithub\ Project Page: \url{https://github.com/ModalityDance/IVT-LR}\par}%
}

% ============================================================
% Tiease figure (if needed)
% ============================================================

% \begin{teaserfigure}
%   \includegraphics[width=\textwidth]{figures/sampleteaser.pdf}
%   \caption{An example of a tiease figure. Delete if you don’t need it.}
%   \label{fig:teaser}
% \end{teaserfigure}

\maketitle

% ============================================================
% Main Sections (add more if needed)
% ============================================================
\section{Introduction}

Over the past few years, the capabilities of large language models (LLMs) have been further unlocked through advancements in reasoning. Researchers have sought to enhance the reasoning abilities of LLMs, initially through prompting techniques such as Chain-of-Thought prompting ~\cite{wei2022chain}, and more recently by developing large reasoning models using reinforcement learning, like GPT-4o ~\cite{hurst2024gpt} and Deepseek-R1 ~\cite{guo2025deepseek}. Building on the success of reasoning in LLMs, there has been growing interest in the research community to extend these reasoning capabilities to multimodal LLMs (MLLMs). This has brought the promising topic of multimodal reasoning, aiming to improve the performance of models on multimodal tasks, such as VQA, through reasoning.

\begin{figure}[t]
    \centering
    \includegraphics[width=1\linewidth]{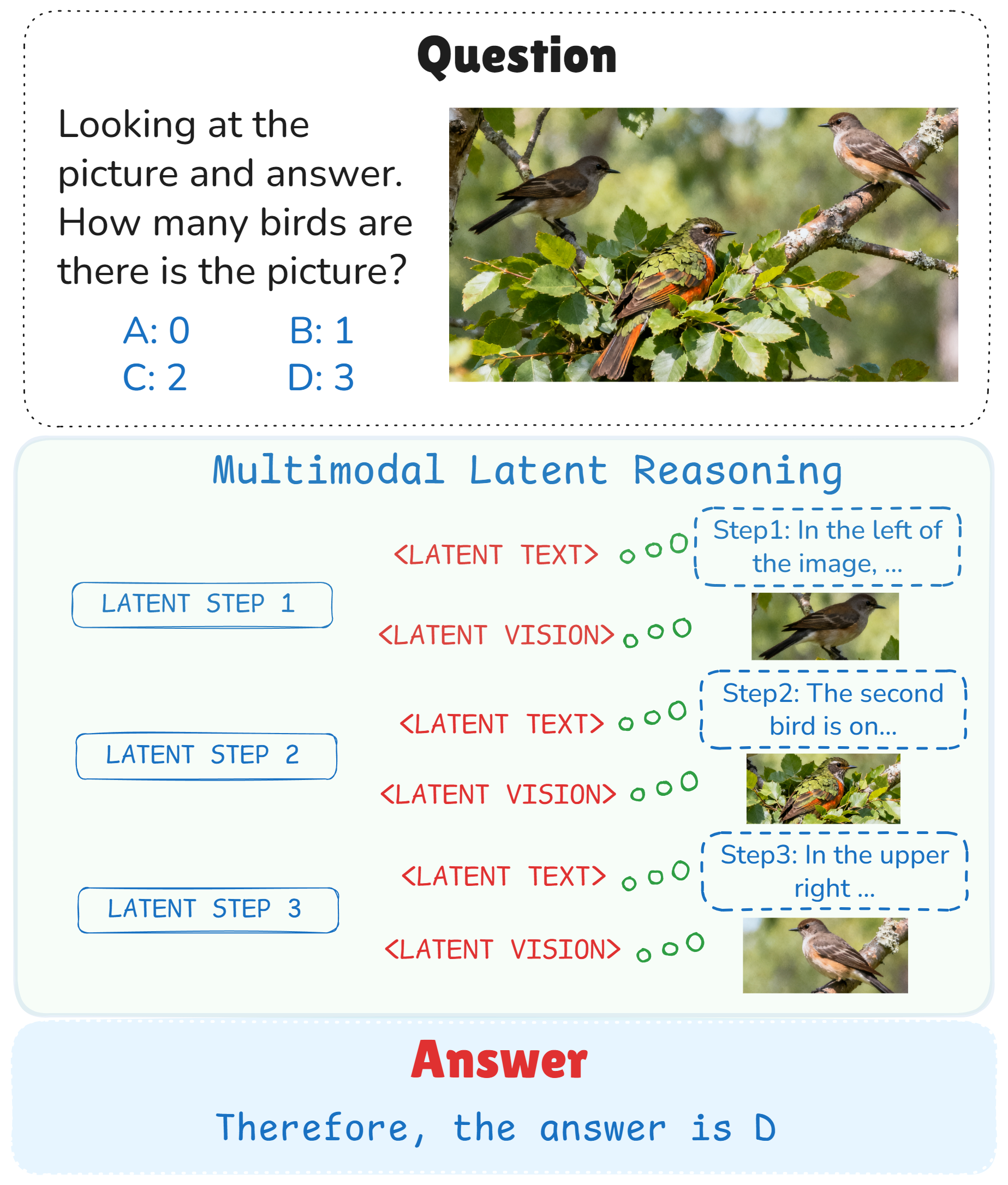}
\caption{An example of interleaved vision-text latent reasoning, where the intermediate reasoning steps are carried out entirely within the multimodal latent space.}
    \label{fig:intro}
\end{figure}

Current multimodal reasoning approaches can be broadly categorized into the following progressive steps: 1) Text-only reasoning. Early multimodal reasoning methods primarily focused on pure text-based reasoning, where MLLMs perform textual reasoning before generating the final answer. These approaches \cite{hu2022promptcap, kamcot} could seamlessly apply LLMs methodologies to MLLMs. 2) Vision-text involved reasoning. Some studies ~\cite{visualsketchpad, liu2025Visual, chern2025thinking} highlight that the intermediate reasoning steps also require the involvement of visual information. For instance,  \citet{gao2024interleaved} enhances reasoning by generating sequential steps that interleave visual information with textual rationales via selected image patches. In a related direction, \citet{zheng2025deepeyes} trains models through end-to-end reinforcement learning to autonomously zoom in on image regions for fine-grained visual inspection during the reasoning process. Alternatively, \citet{li2025imaginereasoningspacemultimodal} enables MLLMs to actively “think visually” by generating explicit image visualizations of their reasoning traces, thereby significantly enhancing performance on complex spatial reasoning tasks.

Recently, latent reasoning has emerged as a new paradigm in LLMs, which eliminates the need for explicit and lengthy textual reasoning by leveraging implicit latent vectors \cite{hao2024training}. Inspired by this, we believe that latent reasoning holds even greater potential for facilitating vision-text interleaved intermediate reasoning steps due to the following reasons: 1) Multimodal representation potential. Latent reasoning enables the reasoning process to occur entirely within a hidden space, offering a greater capacity to represent rich, multimodal information during reasoning. 2) Reduced Annotation. Introducing latent reasoning will lessen reliance on heavily annotated vision-text interleaved reasoning data, as reasoning steps no longer need to be fully observable or linguistically aligned. 3) Inference efficiency. By avoiding long chains of explicit multimodal representation in the reasoning step, it will significantly improve efficiency.

In this work, we propose the Interleaved Vision-Text Latent Reasoning (IVT-LR) method, which enables both textual and visual modalities to perform reasoning entirely in latent space. As shown in Figure~\ref{fig:intro}, in our framework, each latent reasoning step consists of two parts: \emph{latent text} and \emph{latent vision}. At each reasoning step, we use the hidden state from the previous step to replace explicit text as the \emph{latent text} component. Afterwards, for \emph{latent vision} part, a certain number of image embeddings are selected based on their attention scores then concatenated with the hidden state to serve as input for the subsequent reasoning step. To effectively blend the \emph{latent text} and \emph{latent vision} components for joint reasoning in the latent space, we introduce a progressive, multi-stage training strategy that gradually substitutes explicit CoT steps with latent reasoning steps, where supervision is focused on the remaining future steps and the final answer to ensure accurate inference.

The key contributions are summarized: 
\begin{itemize}
    \item We introduce IVT-LR, the first framework to achieve fully unified multimodal latent reasoning. Unlike prior methods, our approach enables both textual and visual information to be reasoned with in the latent space, eliminating the need for intermediate explicit text or image generation.

    \item Our method presents a novel training paradigm that is both data-efficient and computationally efficient, without requiring explicit annotations for intermediate visual reasoning steps. By reasoning in latent space, it also drastically reduces the number of autoregressive steps required for inference.

    \item We validate the effectiveness of IVT-LR through extensive experiments on challenging visual question answering benchmarks, including M$^3$COT and ScienceQA, where our model establishes new state-of-the-art performance in accuracy and significantly improves inference efficiency, as measured by fewer autoregressive steps and lower inference latency.

\end{itemize}
\section{Related Work}

\subsection{Multimodal Reasoning}

Multimodal reasoning focuses on enabling models to reason over information from different modalities to solve complex tasks. Existing approaches can be roughly divided into text-only reasoning and interleaved reasoning.

\paragraph{Text-only reasoning.} Early works attempt to convert visual information into text before reasoning, using tools or visual experts to generate textual representations to guide LLMs. \citet{hu2022promptcap} first introduced the concept of captions, extracting visual content as textual captions and concatenating them to the input to enhance reasoning. Inspired by this, subsequent works pursued a fine-grained understanding of images to improve textual expressiveness. \citet{Zheng_NeurIPS2023} generates a rationale that incorporates image information from visual-text inputs, which is then used for reasoning. Other works~\cite{MitraCCoT, kamcot} leverage graph structures to identify entities in images and construct relationships among them, enhancing reasoning based on these inter-entity connections.

\paragraph{Vision-text involved reasoning.} This line of work emphasizes using images together with text during the rationale generation and reasoning process. Building on the reasoning paradigm of large language models, \citet{zhang2023multimodal} first proposed decoupling rationale generation from answer generation in the Vision-Text Reasoning field. Subsequently, \citet{Shao2024VisualCA} annotates key regions of the original image in intermediate steps, training models to focus on image regions relevant to the answer. While some works~\cite{gao2024interleaved, zhang2025chain} further extract key image regions progressively during reasoning, combining visual information with textual reasoning to generate the final answer. Moreover, new methods~\cite{visualsketchpad, liu2025Visual} emulate human thought by sketching images during reasoning, focusing on core concepts, structures, and relationships while ignoring redundant details. Other works~\cite{li2025imaginereasoningspacemultimodal, chern2025thinking} generate new images during reasoning, combining them with text to improve reasoning in complex scenarios. To completely decouple reasoning from language and amplify the role of images, \citet{xu2025visualplanningletsthink} proposes reasoning solely with newly generated images, achieving substantial improvements in visual navigation tasks.

\subsection{Latent Reasoning}

Latent reasoning refers to internal, non-linguistic thinking performed in a hidden latent space before generating the final answer. Early methods used special tokens to guide latent reasoning. \citet{goyal2023think} introduces learnable \texttt{<pause>} tokens, giving the model opportunities to internally update information before generating an answer, while \citet{wang2023guiding} uses \texttt{<plan>} tokens to guide reasoning.

Later, some works exploit the model's continuous hidden states to replace explicit reasoning steps. \citet{hao2024training} pioneers continuous latent space reasoning by feeding the last hidden states as input embeddings for the next step without generating intermediate tokens, significantly reducing reasoning tokens and improving efficiency. Inspired by this, subsequent methods improve the quality of intermediate representations. \citet{cheng2024compressed} uses variable-length contemplation tokens for latent reasoning, addressing quality degradation caused by fixed-length embeddings. \citet{shen2025codi} employs self-distillation to align student and teacher hidden activations under CoT supervision, constraining latent reasoning paths.  

In the multimodal domain, latent reasoning has also been introduced. Unlike traditional LLMs, VLMs emphasize how image features interact with the latent space. Some efforts \cite{yang2025machine, li2025latent, pham2025multimodal} have been made to integrate visual "thoughts" into the latent space for reasoning. However, these existing works focus solely on single-modal latent reasoning. Combining text and vision for multimodal latent reasoning in the latent space remains unexplored.
\section{Method}

\begin{figure*}[t]
    \centering
    \includegraphics[width=1.0\linewidth]{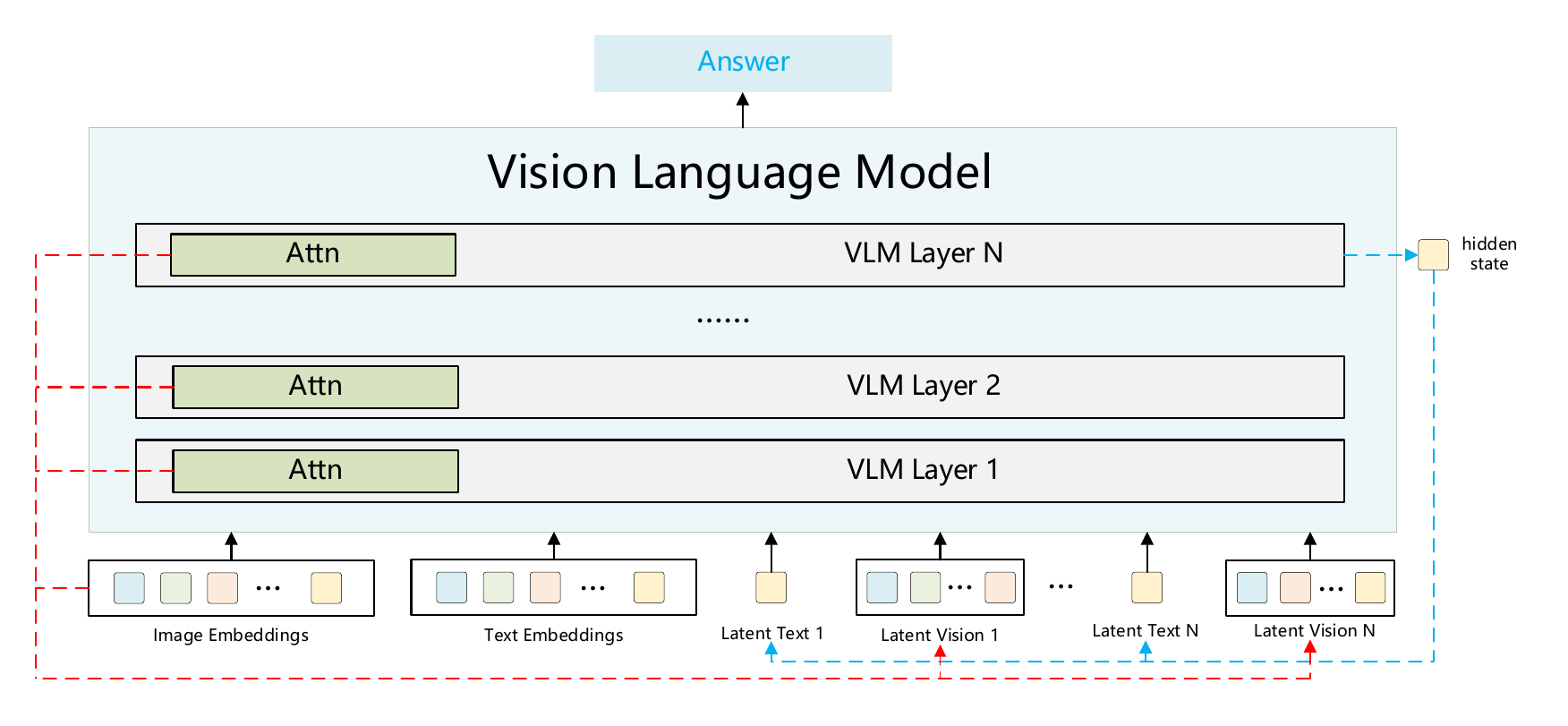}
    \caption{Overview of our Interleaved Vision-Text Latent Reasoning (IVT-LR) framework. At each step, reasoning is performed entirely in the latent space by fusing $\emph{latent text}$ (the hidden state from the previous step) and $\emph{latent vision}$ (dynamically selected image embeddings based on attention scores).}
    \label{fig:method}
\end{figure*}

In this section, we present IVT-LR, the first VLM framework that unifies textual and visual representations in the latent space and implements multimodal latent reasoning. Given a text sequence $\mathcal{X} = (x_1, \dots, x_I)$ and a set of visual embeddings $\mathcal{Z} = (z_1, \dots, z_J)$ from a visual encoder, a standard VLM encodes the text sequence into embeddings, incorporates visual features, and predicts a conditional distribution over the next token:
\[
e_{1:t}^{\text{text}} = g(x_{1:t}) \in \mathbb{R}^{t \times d},
\]
\[
e_t^{\text{fused}} = f(e_{1:t}^{\text{text}}, \mathcal{Z}) \in \mathbb{R}^d,
\]
\[
\mathcal{M}(x_{t+1} \mid x_{1:t}, \mathcal{Z}) = \text{softmax}(W \cdot e_t^{\text{fused}}) ,
\]
where $g(\cdot)$ denotes the text embedding function, $f(\cdot)$ is a function that generates the hidden state for the next token based on the textual embeddings and visual fatures, and $W \in \mathbb{R}^{|V| \times d}$ is trained to project the fused representation to a distribution over the vocabulary. This formulation illustrates how a VLM predicts the next token conditioned on both textual context and visual information.

\begin{figure*}[t]
    \centering
    \includegraphics[width=1\linewidth]{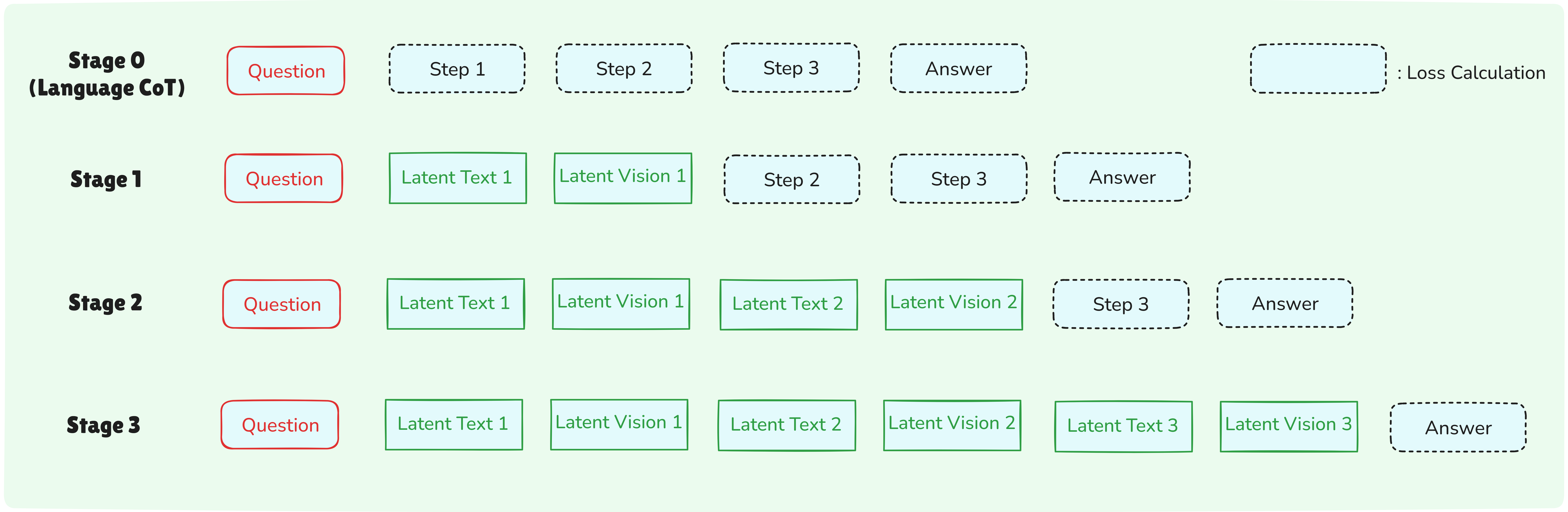}
    \caption{Overview of the Multi-Stage Progressive Training Strategy used for IVT-LR. The strategy begins with full explicit CoT and then gradually substitutes one explicit reasoning step with latent text and latent vision. Training loss is calculated exclusively over the remaining explicit steps and the final answer.}
    \label{fig:training}
\end{figure*}

\subsection{Multimodal Latent Reasoning}
Figure~\ref{fig:method} provides an overview of our approach. In IVT-LR, the latent reasoning is conducted over both \emph{latent text} and \emph{latent vision}. Following~\cite{hao2024training}, the textual modality bypasses explicit token prediction: instead of using the embedding of the previous explicit text token, we represent the latent text with the hidden state $h_{t-1}^{hidden}$, which effectively encodes the necessary reasoning logic and preserves richer intermediate information in a continuous latent space. Meanwhile, the latent vision is designed to model the dynamic focus on the visual features at each step. Specifically, we extend latent reasoning to visual modality by selecting the $k$ most relevant visual features from the image embedding set. Thus, an attention-based selection mechanism is designed to choose a fixed number of image embeddings from the full set $[z_1, z_2, \dots, z_J]$. We utilize the sum of attention weights across all layers to identify the $k$ image embedding positions with the highest cumulative scores. The selected features are appended to the hidden states $h_{t-1}^{hidden}$, resulting in a multimodal latent representation $[h_{t-1}^\text{latent}, z_{t-1}^\text{selected}]$.  The input to the model at step $t$ thus consists of all prior hidden states and their selected visual features, along with any preceding question embeddings, which can be written as $E_t = [e_1, \dots, e_N, h_{1}^\text{latent}, z_{1}^\text{selected}, \dots, h_{t-1}^\text{latent}, z_{t-1}^\text{selected}]$. The model fuses these multimodal representations to obtain $e_t^\text{fused} = f(E_t)$, which is projected through the output head to yield the next-token distribution $\mathcal{M}(x_{t+1} \mid E_t) = \text{softmax}(W \cdot e_t^\text{fused})$.  This design allows the model to perform step-wise multimodal latent reasoning without generating intermediate reasoning sequences.

\newcommand{\algcomment}[1]{\STATE \textcolor{blue}{#1}}
\begin{algorithm}[t!]
\caption{IVT-LR}
\begin{algorithmic}[1]
    \STATE \textbf{Input:} Text input embeddings $\mathcal{E}=[e_1,\dots,e_I]$, Image input embeddings $\mathcal{Z} = [z_1,\dots,z_J]$, Whole input embeddings $\mathcal{Q} = \mathcal{E}+\mathcal{Z}=[q_1,\dots,q_m]$, Latent step positions $\mathcal{L} = [l_1,\dots,l_N] $, Number of selected embeddings $k$
    
    \FOR{$i=1$ to $N$}
            \STATE $h_i \gets \text{LastHiddenState}(q_{1:l_i-1})$
            \STATE $\mathcal{Z}_{sel} \gets \text{AttentionSelect}(\mathcal{Z}, k)$
            \STATE $latent[i] \gets [h_i, \mathcal{Z}_{sel}]$
            \STATE $\mathcal{Q}\gets \text{Concat}(\mathcal{Q}, latent[i])$      
            \STATE $l_n \gets l_n + (k+1) \quad \text{for } n > i$
        
    \ENDFOR{}

    \STATE $q_{:end} \gets \text{PredictToEnd}(q_{:l_N})$
    \STATE \textbf{Answer} $\gets \text{Decode}(q_{l_N+1:})$
    \STATE \textbf{return Answer}
\end{algorithmic}
\label{alg:mcoconut}
\end{algorithm}

\subsection{Training Procedure.} The objective of IVT-LR is to enable multimodal reasoning within the latent space. Inspired by ~\citet{deng2024implicitcot}, we adopt a multi-stage training strategy to progressively boost the model's reasoning capability. In the preprocessing stage, each reasoning trajectory is segmented into up to $N$ steps, followed by the final answer. At stage 0, as shown in Figure~\ref{fig:training}, the model is trained with standard CoT supervision, where all reasoning steps are explicitly generated to strengthen symbolic reasoning ability. Afterwards, the latent reasoning steps are progressively introduced within the $N$ stages: at each stage, one additional explicit reasoning step is replaced by a latent reasoning step, denoted by the special token \textit{<latent>}, beginning with the first step. In this way, the model learns to progressively substitute explicit reasoning using latent textual and visual representations while still being supervised on the final answer.

Training is optimized using negative log-likelihood (NLL) loss, with supervision applied only to reasoning steps and the final answer. Latent reasoning steps and question tokens are masked out. This design ensures that the supervision signal is placed only on the reasoning steps and the final answer, distinguishing our approach from standard knowledge distillation. Unlike distillation, which enforces alignment between student hidden states and a teacher’s explicit reasoning trajectory, our method avoids imposing a strict linear path. By avoiding excessive alignment between latent representations and explicit rationales, the model learns to internalize reasoning trajectories in latent space with essential image features, while still being driven toward correct final predictions. Compared to single-step fine-tuning, the proposed multi-stage training introduces additional stages but reuses the same backbone and training data, resulting in a moderate increase in training cost that remains practical and acceptable.

\begin{table*}[t]
\centering
\resizebox{\textwidth}{!}{
\begin{tabular}{cccccccc} % 新增一列，共 8 列
\toprule
\multirow{2}{*}{Backbone} & \multirow{2}{*}{Methods} & \multicolumn{3}{c}{M$^3$CoT} & \multicolumn{3}{c}{ScienceQA} \\
\cmidrule(lr){3-5} \cmidrule(lr){6-8} % 列指标也相应调整
& & Acc.(\%) $\uparrow$ & \# AR Steps $\downarrow$& Avg. Time(s) $\downarrow$& Acc.(\%) $\uparrow$ & \# AR Steps $\downarrow$& Avg. Time(s) $\downarrow$\\
\midrule
\multirow{7}{*}{Qwen2-VL} & No-CoT & 45.4& -& -& 64.4& -& -\\
& Multimodal CoT\citep{zhang2023multimodal} & 42.5& 106.3& 3.10& 58.3& 83.9& 2.44\\
& CCoT\citep{MitraCCoT} & 44.1& 177.2& 5.31& 63.8& 164.0& 5.23\\
& ICoT\citep{gao2024interleaved} & 46.0& 96.5& 2.86& 65.4& 77.4& 2.28\\
& SCAFFOLD\citep{lei2024scaffolding} & 44.9& 170.8& 5.14& 62.5& 162.3& 4.91\\
& Chain-of-Focus\citep{zhang2025chain} & 64.3& 185.7& 2.63& 91.2& 162.3& 2.09\\
& \textbf{IVT-LR} & \textbf{71.8} & \textbf{10.0}& \textbf{0.65}& \textbf{94.6}& \textbf{11.0}& \textbf{0.67}\\
\midrule
\multirow{7}{*}{Chameleon} & No-CoT & 28.4& -& -& 48.5& -& -\\
& Multimodal CoT\citep{zhang2023multimodal} & 30.6& 110.5& 3.62& 50.7& 98.7& 3.33\\
& CCoT\citep{MitraCCoT} & 31.4& 168.4& 5.35& 51.3 & 174.2& 5.39\\
& ICoT\citep{gao2024interleaved} & 32.3& 110.9& 5.43& 53.4& 92.4& 4.62\\
& SCAFFOLD\citep{lei2024scaffolding} & 31.1& 194.3& 6.12& 47.5& 160.6& 6.03\\
& Chain-of-Focus\citep{zhang2025chain} & 36.5& 739.4& 3.09& 61.2& 717.1& 2.56\\
& \textbf{IVT-LR} & \textbf{41.8}& \textbf{10.0}& \textbf{1.13}& \textbf{64.0} & \textbf{11.0}& \textbf{1.14}\\
\bottomrule
\end{tabular}}
\captionsetup{skip=4pt}
\caption{Comparison of IVT-LR with various multimodal reasoning baselines on the M$^3$CoT and ScienceQA benchmarks. The reported metrics include: Answer Accuracy (Acc.), Average number of Autoregressive Steps (\# AR Steps), and Average Generation Time (Avg. Time). Experiments are conducted using two backbone models: Qwen2-VL-7B and Chameleon-7B.}
\label{tab:comparison}
\end{table*}

\subsection{Inference Process.} Since all rationales in training have been segmented into a certain number of steps, at inference time, the same number of \textit{<latent>} tokens are appended after the question and image inputs. This setup ensures that reasoning is fully conducted in latent space and no explicit reasoning steps are produced before the final answer. 

To evaluate the intermediate models at stage $n$, inference uses $n$ latent tokens, yielding mixed explicit-latent reasoning consistent with the training stage. Importantly, latent text and latent vision co-exist only during the latent reasoning phase, where visual evidence is integrated into the hidden trajectory. Outside this phase, the model operates in a purely linguistic generation mode. 

\section{Experiments}

\subsection{Experimental Setup.} 
\paragraph{Datasets and Evaluation.}
We evaluate our method on two widely used multimodal reasoning benchmarks: M$^3$CoT \citep{chen-etal-2024-m3cot} and ScienceQA \citep{lu2022learn}. M$^3$CoT is a large-scale benchmark focusing on multimodal chain-of-thought reasoning, where models must combine both visual and textual inputs to perform multi-step reasoning. ScienceQA is a diverse dataset covering natural science, language science, and social science, with many questions accompanied by diagrams or images.
We evaluate using exact-match answer accuracy, along with the average number of autoregressive steps and the average response time per question. These metrics capture both correctness and reasoning efficiency.

\paragraph{Baselines and Implementation Details}

We compare IVT-LR against six representative methods, including text-only reasoning: CCoT~\citep{MitraCCoT}; vision-text involved reasoning: Chain-of-Focus~\citep{zhang2025chain}, SCAFFOLD~\citep{lei2024scaffolding}, ICoT~\citep{gao2024interleaved}, Multimodal-CoT~\citep{zhang2023multimodal}; and No-CoT that directly predicts answers without generating intermediate steps. 

For fair comparison, we evaluate IVT-LR and all baselines with Qwen2-VL-7B~\citep{wang2024qwen2} and Chameleon-7B~\citep{team2024chameleon} backbones. In IVT-LR training, we use a stage number ($N$) of four (detailed discussion provided in Appendix \ref{sec:appendix}) , a batch size of four, and train with the Adam optimizer where the learning rate is set to $\mathbf{4 \times 10^{-5}}$ and $\boldsymbol{\beta}_1$ is set to $0.9$. All experiments run on four NVIDIA A6000 GPUs (48GB VRAM each).

\subsection{Main Results.} 
The results on M$^3$CoT and ScienceQA are summarized in Table~\ref{tab:comparison}. Analyzing these outcomes, we draw the following key observations:

\paragraph{Multimodal Reasoning Accuracy.} 
IVT-LR achieves the highest accuracy on both the M$^3$CoT and ScienceQA benchmarks, consistently outperforming all baselines with both Qwen2-VL and Chameleon backbones. Compared to the strongest baseline, Chain-of-Focus, IVT-LR yields improvements of 5\% (Chameleon backbone) to 7.5\% (Qwen2-VL backbone) on M$^3$CoT. Similar gains are observed on the ScienceQA benchmark. Beyond this, IVT-LR surpasses other methods by margins of 10\% to 25\%, depending on the backbone and task. These results demonstrate that IVT-LR enables more effective cross-modal interaction in the latent space, leading to stronger multimodal reasoning capability on complex tasks.

\paragraph{Reasoning Efficiency.} 
Beyond accuracy, a critical advantage of IVT-LR is its significantly enhanced inference efficiency, which is quantified by fewer autoregressive steps and lower inference latency compared to baselines. 1) Fewer autoregressive steps. Across both backbones, IVT-LR achieves at least a $\mathbf{9\times}$ reduction in the number of autoregressive steps required for generation compared to most baselines. This efficiency is achieved by conducting reasoning in the latent space, avoiding the need for lengthy, explicitly generated rationales required by other methods. 2) Lower Inference Latency. With the Qwen model, IVT-LR achieves an average inference time of approximately 0.66s, making it 3 to 8 times faster than all other baselines. A similar trend of significant speedup holds true for the Chameleon backbone. While No-CoT achieves the absolute lowest latency by completely sacrificing deep reasoning (around 0.35s), IVT-LR delivers state-of-the-art accuracy at an inference speed only marginally longer than the minimal No-CoT, demonstrating superior efficiency in the high-accuracy setting. 

In summary, IVT-LR demonstrates both superior accuracy and improved reasoning efficiency in VQA tasks. By performing multi-step reasoning in latent space, the model not only achieves the highest accuracy among all baselines but also significantly reduces the number of autoregressive steps and achieves a substantially lower inference latency. These results highlight the effectiveness of latent reasoning in combining textual and visual info.

\subsection{Ablation Study.} 

\begin{table}[t!]
  \centering
  \renewcommand{\arraystretch}{1.2} % 调整行高
  \resizebox{0.98\linewidth}{!}{
    \begin{tabular}{c|cc}
    \bottomrule
    Methods & M$^3$CoT & ScienceQA \\
    \hline
     IVT-LR & \textbf{71.83}& \textbf{94.1}\\
     w/o latent text & 52.20 (\textcolor{red}{-19.63}) & 84.7 (\textcolor{red}{-9.8})\\
     w/o latent vision & 46.64 (\textcolor{red}{-25.19}) & 82.3 (\textcolor{red}{-11.8})\\
     w/o the whole latent part & 58.02 (\textcolor{red}{-13.81})& 86.4 (\textcolor{red}{-7.7})\\
    \toprule
    \end{tabular}}
  \caption{
Accuracy comparison of IVT-LR on Qwen2-VL, showing the performance impact with and without its core latent components ($\emph{latent text}$ and/or $\emph{latent vision}$). Values in parentheses indicate performance drop relative to full IVT-LR.
  }
  \label{tab:io}%
\end{table}%

To verify the necessity of IVT-LR’s two key components, latent text and latent vision, we conducted a series of ablation experiments on visual reasoning tasks. Specifically, we evaluated the effects of removing latent text, latent vision, and both components simultaneously.

\paragraph{Latent Text.}  
As shown in Table~\ref{tab:io}, removing latent text(w/o latent text) leads to a noticeable drop in accuracy on both M$^3$CoT and ScienceQA. This demonstrates that latent text plays a crucial role in model performance: it provides a compact, continuous representation of intermediate reasoning states. This allows the model to internalize multi-step reasoning trajectories directly in the latent space, avoiding biases introduced by language-based alignment. Furthermore, operating in continuous hidden spaces, it effectively mitigates the amplification of errors typical in discrete, step-by-step textual reasoning.

\paragraph{Latent Vision.}  
Table~\ref{tab:io} also shows that removing latent vision (w/o latent vision) also results in decreased performance. This indicates that incorporating the most informative visual cues is vital for precise multimodal reasoning. Without this mechanism, the model cannot focus on the critical regions of the image, reducing the effectiveness of each reasoning step. Besides, based on attention-driven integration, latent vision ensures that the latent space receives rich, contextually relevant visual information. It also mitigates interference from irrelevant image regions, leading to more accurate and robust reasoning.

\subsection{In-depth Analysis.}

\begin{figure}[t]
    \centering
    % 左子图
    \begin{subfigure}{0.48\columnwidth}
        \centering
        \includegraphics[width=\linewidth]{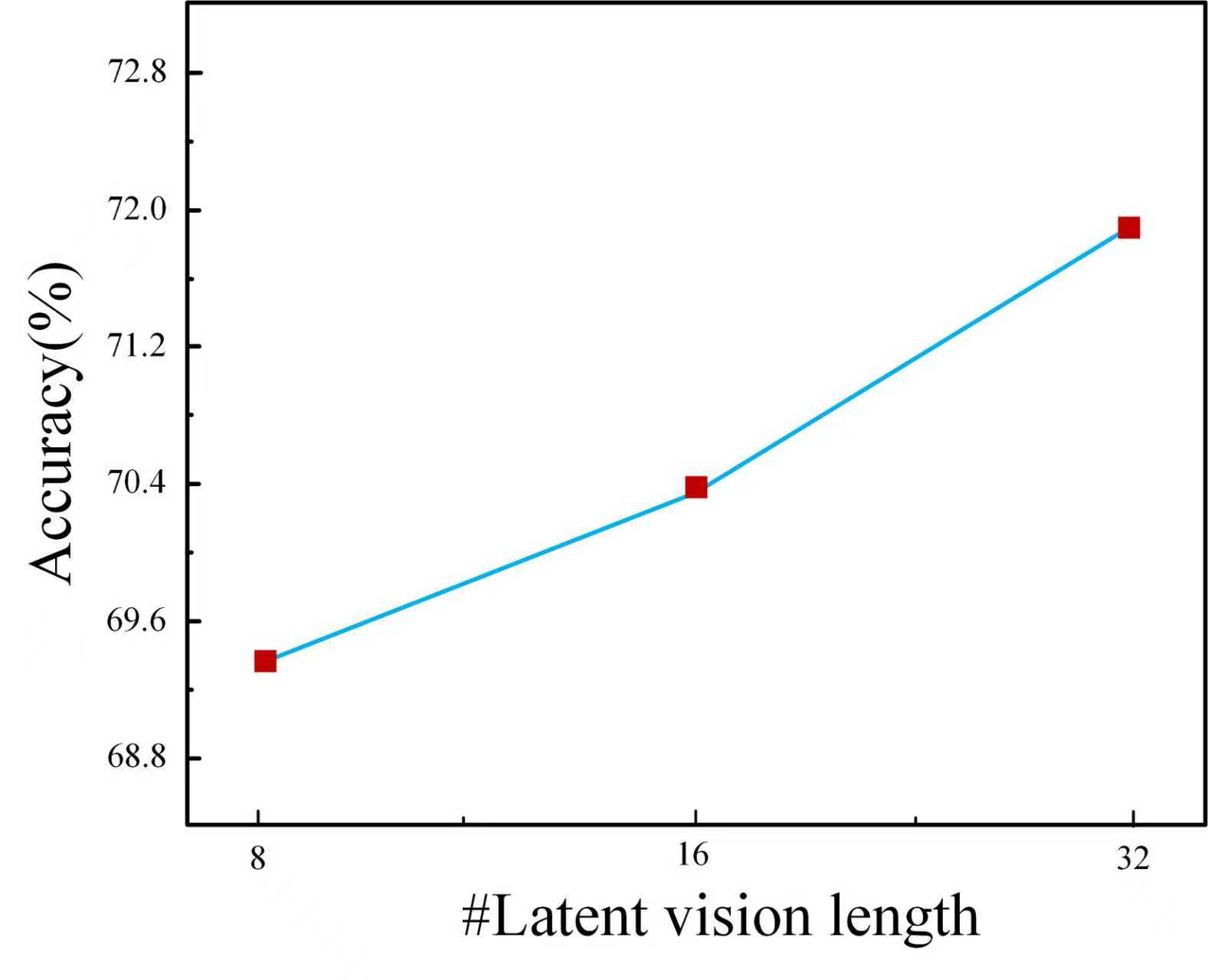}
        \caption{M$^3$CoT}
        \label{fig:left}
    \end{subfigure}
    \hfill
    % 右子图
    \begin{subfigure}{0.48\columnwidth}
        \centering
        \includegraphics[width=\linewidth]{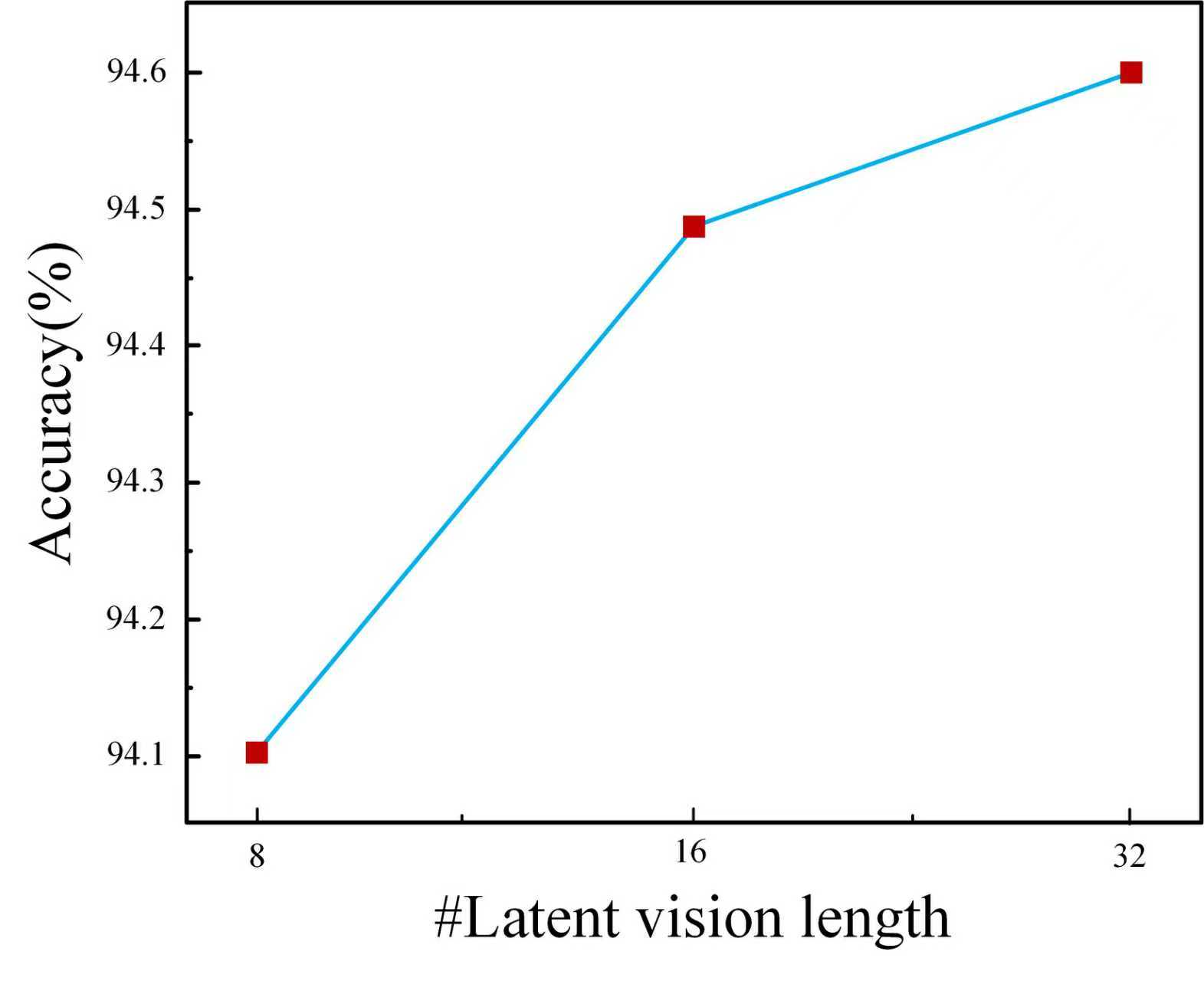}
        \caption{ScienceQA}
        \label{fig:right}
    \end{subfigure}
    
    \caption{Accuracy comparison of IVT-LR on the length of \emph{latent vision} per reasoning step across two reasoning benchmarks: (a) M$^3$CoT and (b) ScienceQA.}
    \label{fig:combined}
\end{figure}

\paragraph{Length of Latent vision}  
We investigate the impact of varying the latent vision length per step.As shown in Figure~\ref{fig:combined}, accuracy steadily increases with this length, indicating that longer latent vision sequences provide richer visual cues necessary for complex reasoning. Since the latent vision is formed by adaptively selecting visual embeddings from the image, increasing the length of these selections allows the model to gradually approach full-image utilization (e.g., 32 embeddings over three steps roughly cover the whole image in Qwen2-VL). This ensures that essential visual details, often required for global comprehension, are not omitted. Moreover, because embeddings are selected step-by-step across the latent reasoning stages, the process achieves targeted and cumulative coverage: each round complements the previous ones, enabling the model to integrate both localized critical features and broader global context in a structured, effective manner.

\paragraph{Stages of Latent Reasoning}  
We evaluate models with $1$, $2$, and $3$ latent reasoning steps to study the effect of progressively replacing explicit reasoning. As shown in Table~\ref{tab:latent_stages}, accuracy improves as more reasoning steps are conducted in latent space, showing that latent representations provide a more robust reasoning mechanism than explicit language. This is because latent states avoid errors from language alignment and allow smoother integration with image embeddings.

Domain-wise results show that science and mathematics benefit most from additional latent tokens, highlighting that structured reasoning tasks are particularly suited for latent-space inference. The accuracy in commonsense also improves, but with smaller gains, since it often relies less on multi-step deduction. Together, these findings confirm that latent reasoning scales effectively with task complexity, supporting both efficiency and accuracy.

\paragraph{Attention Shift over Step-wise Embeddings}
To further investigate the internal mechanisms of IVT-LR, we analyze how the model allocates its attention to image embeddings under our method and explicit reasoning with selected image embeddings. We use \textbf{Attention Ratio} and \textbf{Attention Focus} as metrics to analyze the model's focus.

\begin{table}[t]
  \centering
  \renewcommand{\arraystretch}{1.5} % 调整行高
  \resizebox{0.98\linewidth}{!}{
    \begin{tabular}{c|cccc}
    \hline
    Latent Stage & Science & Commonsense & Mathematics & Total \\
    \hline
    1 & 56.66\% & 64.40\% & 38.59\% & 56.30\% \\
    2 & 61.71\% & 70.11\% & 43.57\% & 61.48\% \\
    3 & \textbf{70.90\%} & \textbf{79.78\%} & \textbf{63.07\%} & \textbf{71.83\%} \\
    \hline
    \end{tabular}}
  \caption{Accuracy on M$^3$CoT across different latent reasoning stages. Results are shown both overall and broken down by domain.}
  \label{tab:latent_stages}
\end{table}

\noindent\textbf{(1) Attention Ratio:}  
\begin{equation}
R = \frac{\sum_{j \in \mathcal{I}} \text{Attn}(E_j)}{\sum_{i \in \mathcal{T}} \text{Attn}(E_i)},
\end{equation}
where $\mathcal{I}$ denotes the visual reasoning part, specifically the set of selected image embeddings, and $\mathcal{T}$ denotes the text tokens or the latent text part. This ratio reflects the relative allocation of attention between visual and textual information.

\noindent\textbf{(2) Attention Focus (Inverse Entropy):}  
\[
H = - \sum_{k} p_k \log p_k, \quad 
p_k = \frac{\text{Attn}(E_k)}{\sum_{m} \text{Attn}(E_m)},
\]
\begin{equation}
F = \frac{1}{H + \epsilon}, \quad \epsilon \ll 1,
\end{equation}
where $F$ is the Attention Focus. Higher $F$ indicates more concentrated attention, while lower $F$ reflects dispersed focus.

\begin{figure}[t]
    \centering
    \includegraphics[width=\linewidth]{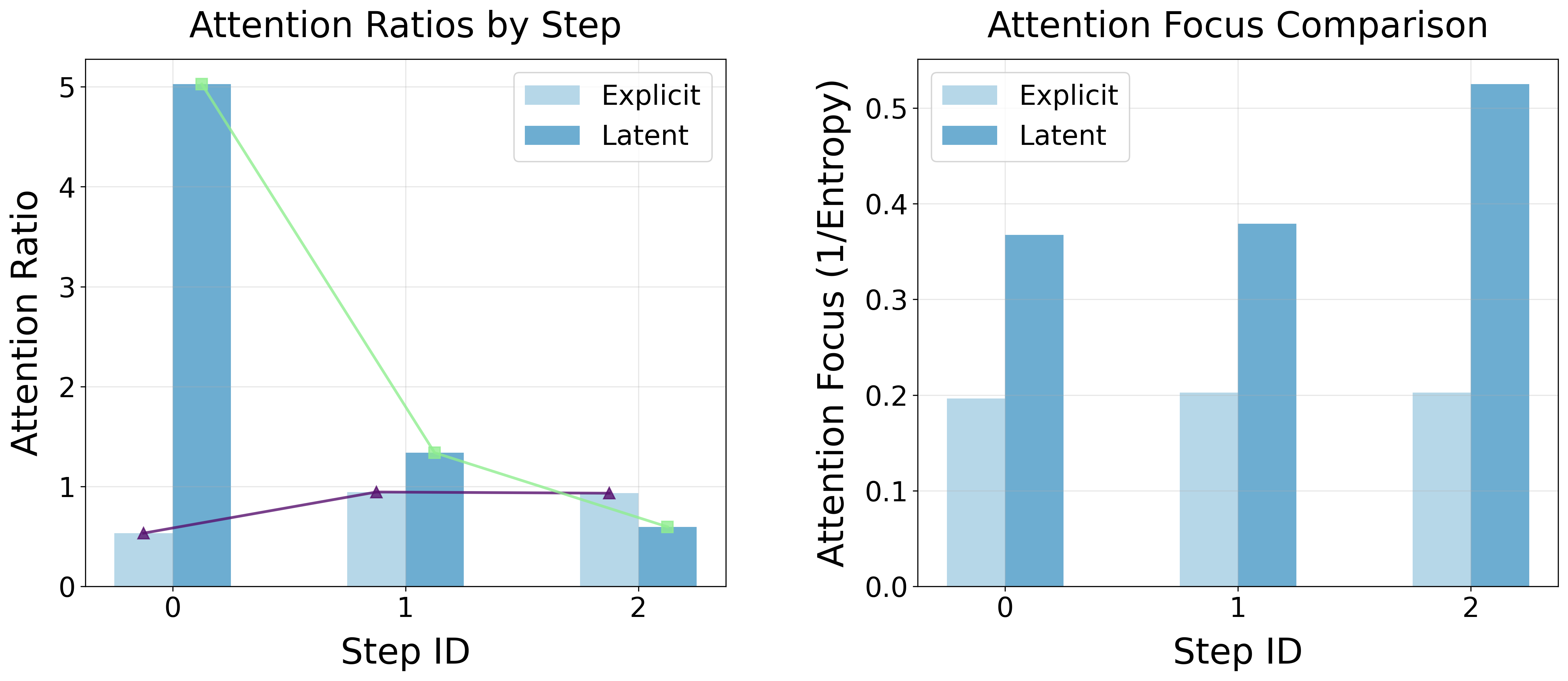}
    \caption{Attention analysis comparison between explicit and latent reasoning approaches. Left: attention ratios of visual part to textual part across reasoning steps. Right: attention focus measured by inverse entropy.}
    \label{fig:attention_ana}
\end{figure}

The result is shown in Figure~\ref{fig:attention_ana}. We found significant differences in model behavior between latent and explicit multimodal reasoning modes: 

\textbf{(1) Dynamic Attention Ratio: A Core of Visio-Linguistic Perception.}  
In the latent reasoning mode, the attention ratio exhibits a clear downward trend across reasoning steps. Initially, the model focuses predominantly on latent vision, but over subsequent steps, attention gradually shifts to its latent text for deeper textual reasoning. This dynamic adjustment demonstrates the model’s ability to prioritize the most informative visual cues and adaptively reallocate focus, reflecting enhanced visio-linguistic perception. In contrast, under explicit reasoning, the attention ratio remains largely unchanged and consistently below 1, indicating persistent focus on textual tokens. This suggests that, with interference from abundant textual information, explicit reasoning struggles to effectively filter and leverage critical visual features.

\textbf{(2) Rising Attention Focus: A Hallmark of Efficient Reasoning.}  
Beyond changes in attention ratio, our analysis of attention focus also reveals important insights. In latent reasoning, attention focus shows a progressively increasing trend, showing that the model’s attention becomes increasingly concentrated over reasoning steps. This suggests that at each step, the model effectively filters and refines multimodal information, gradually converging on the most critical and relevant cues—a pattern reminiscent of human problem-solving, where distractions are progressively eliminated and attention is concentrated on core evidence. In contrast, under explicit reasoning, attention focus is not only markedly lower than in implicit reasoning but also exhibits little change across steps. This indicates that explicit reasoning distributes attention more diffusely and lacks clear direction, processing substantial amounts of redundant or less relevant information, which reduces reasoning efficiency and limits the effective extraction of key visual-textual information.

\section{Conclusion}

In this work, we present IVT-LR, the first vision-language reasoning framework that performs multimodal latent reasoning. IVT-LR utilizes latent text and latent vision to internalize complex reasoning trajectories, thereby realizing comprehensive multimodal latent reasoning. This approach effectively mitigates the attention dilution problem present in existing methods that rely on explicit textual reasoning and full-image processing. On VQA and other visual reasoning tasks, IVT-LR significantly outperforms multiple strong baselines, achieving new state-of-the-art results in both reasoning accuracy and efficiency. Our findings demonstrate the potential of interleaved vision-text reasoning in latent space, offering a promising paradigm for building more efficient and perceptive vision-language models and inspiring future research on multimodal reasoning strategies.

Future work could explore more dynamic ways of visual latent reasoning, such as adaptively determining the optimal number of latent steps based on the complexity of the question, rather than relying on a fixed stage number. Furthermore, this approach is highly promising for extending its application beyond pure reasoning to broader sequential multimodal tasks, including planning and complex decision-making in dynamic environments.

% ============================================================
% Bibliography
% ============================================================

\bibliographystyle{unsrtnat} 
\bibliography{ref}

@String(AAAI = {AAAI})

@article{wei2022chain,
  title={Chain-of-thought prompting elicits reasoning in large language models},
  author={Wei, Jason and Wang, Xuezhi and Schuurmans, Dale and Bosma, Maarten and Xia, Fei and Chi, Ed and Le, Quoc V and Zhou, Denny and others},
  journal={Advances in neural information processing systems},
  volume={35},
  pages={24824--24837},
  year={2022}
}

@article{guo2025deepseek,
  title={Deepseek-r1: Incentivizing reasoning capability in llms via reinforcement learning},
  author={Guo, Daya and Yang, Dejian and Zhang, Haowei and Song, Junxiao and Zhang, Ruoyu and Xu, Runxin and Zhu, Qihao and Ma, Shirong and Wang, Peiyi and Bi, Xiao and others},
  journal={arXiv preprint arXiv:2501.12948},
  year={2025}
}

@inproceedings{gao2024interleaved,
  title={Interleaved-modal chain-of-thought},
  author={Gao, Jun and Li, Yongqi and Cao, Ziqiang and Li, Wenjie},
  booktitle={Proceedings of the Computer Vision and Pattern Recognition Conference},
  pages={19520--19529},
  year={2025},
  doi={10.1109/CVPR52734.2025.01818}
}

@article{deng2024implicitcot,
  title={From explicit cot to implicit cot: Learning to internalize cot step by step},
  author={Deng, Yuntian and Choi, Yejin and Shieber, Stuart},
  journal={arXiv preprint arXiv:2405.14838},
  year={2024}
}

@article{zhang2025chain,
  title={Chain-of-Focus: Adaptive Visual Search and Zooming for Multimodal Reasoning via RL},
  author={Zhang, Xintong and Gao, Zhi and Zhang, Bofei and Li, Pengxiang and Zhang, Xiaowen and Liu, Yang and Yuan, Tao and Wu, Yuwei and Jia, Yunde and Zhu, Song-Chun and others},
  journal={arXiv preprint arXiv:2505.15436},
  year={2025}
}

@article{zhang2023multimodal,
  title={Multimodal Chain-of-Thought Reasoning in Language Models},
  author={Zhang, Zhuosheng and Zhang, Aston and Li, Mu and Zhao, Hai and Karypis, George and Smola, Alex},
  journal={Transactions on Machine Learning Research},
  volume={2024},
  year={2024},
  publisher={Transactions on Machine Learning Research}
}

@inproceedings{chen-etal-2024-m3cot,
  title={M3CoT: A Novel Benchmark for Multi-Domain Multi-step Multi-modal Chain-of-Thought},
  author={Chen, Qiguang and Qin, Libo and Zhang, Jin and Chen, Zhi and Xu, Xiao and Che, Wanxiang},
  booktitle={Proceedings of the 62nd Annual Meeting of the Association for Computational Linguistics (Volume 1: Long Papers)},
  pages={8199--8221},
  year={2024},
  doi={10.18653/v1/2024.acl-long.446}
}

@inproceedings{lu2022learn,
  title={Learn to explain: multimodal reasoning via thought chains for science question answering},
  author={Lu, Pan and Mishra, Swaroop and Xia, Tony and Qiu, Liang and Chang, Kai-Wei and Zhu, Song-Chun and Tafjord, Oyvind and Clark, Peter and Kalyan, Ashwin},
  booktitle={Proceedings of the 36th International Conference on Neural Information Processing Systems},
  pages={2507--2521},
  year={2022}
}

@article{hao2024training,
  title={Training large language models to reason in a continuous latent space},
  author={Hao, Shibo and Sukhbaatar, Sainbayar and Su, DiJia and Li, Xian and Hu, Zhiting and Weston, Jason and Tian, Yuandong},
  journal={arXiv preprint arXiv:2412.06769},
  year={2024}
}

@inproceedings{MitraCCoT,
  title={Compositional chain-of-thought prompting for large multimodal models},
  author={Mitra, Chancharik and Huang, Brandon and Darrell, Trevor and Herzig, Roei},
  booktitle={Proceedings of the IEEE/CVF Conference on Computer Vision and Pattern Recognition},
  pages={14420--14431},
  year={2024},
  doi={10.1109/CVPR52733.2024.01367}
}

@inproceedings{lei2024scaffolding,
  title={Scaffolding Coordinates to Promote Vision-Language Coordination in Large Multi-Modal Models},
  author={Lei, Xuanyu and Yang, Zonghan and Chen, Xinrui and Li, Peng and Liu, Yang},
  booktitle={Proceedings of the 31st International Conference on Computational Linguistics},
  pages={2886--2903},
  year={2025}
}

@article{hu2022promptcap,
  title={Promptcap: Prompt-guided task-aware image captioning},
  author={Hu, Yushi and Hua, Hang and Yang, Zhengyuan and Shi, Weijia and Smith, Noah A and Luo, Jiebo},
  journal={arXiv preprint arXiv:2211.09699},
  year={2022}
}

@inproceedings{Zheng_NeurIPS2023,
  title={DDCoT: duty-distinct chain-of-thought prompting for multimodal reasoning in language models},
  author={Zheng, Ge and Yang, Bin and Tang, Jiajin and Zhou, Hong-Yu and Yang, Sibei},
  booktitle={Proceedings of the 37th International Conference on Neural Information Processing Systems},
  pages={5168--5191},
  year={2023}
}

@inproceedings{kamcot,
  title={Kam-cot: Knowledge augmented multimodal chain-of-thoughts reasoning},
  author={Mondal, Debjyoti and Modi, Suraj and Panda, Subhadarshi and Singh, Rituraj and Rao, Godawari Sudhakar},
  booktitle={Proceedings of the AAAI conference on artificial intelligence},
  volume={38},
  number={17},
  pages={18798--18806},
  year={2024},
  doi={10.1609/aaai.v38i17.29844}
}

@inproceedings{visualsketchpad,
  title={Visual SKETCHPAD: sketching as a visual chain of thought for multimodal language models},
  author={Hu, Yushi and Shi, Weijia and Fu, Xingyu and Roth, Dan and Ostendorf, Mari and Zettlemoyer, Luke and Smith, Noah A and Krishna, Ranjay},
  booktitle={Proceedings of the 38th International Conference on Neural Information Processing Systems},
  pages={139348--139379},
  year={2024}
}

@article{liu2025Visual,
  title={Visual Abstract Thinking Empowers Multimodal Reasoning},
  author={Liu, Dairu and Wang, Ziyue and Ruan, Minyuan and Luo, Fuwen and Chen, Chi and Li, Peng and Liu, Yang},
  journal={arXiv preprint arXiv:2505.20164},
  year={2025}
}

@inproceedings{Shao2024VisualCA,
  title={Visual CoT: advancing multi-modal language models with a comprehensive dataset and benchmark for chain-of-thought reasoning},
  author={Shao, Hao and Qian, Shengju and Xiao, Han and Song, Guanglu and Zong, Zhuofan and Wang, Letian and Liu, Yu and Li, Hongsheng},
  booktitle={Proceedings of the 38th International Conference on Neural Information Processing Systems},
  pages={8612--8642},
  year={2024}
}

@article{li2025imaginereasoningspacemultimodal,
  title={Imagine while reasoning in space: Multimodal visualization-of-thought},
  author={Li, Chengzu and Wu, Wenshan and Zhang, Huanyu and Xia, Yan and Mao, Shaoguang and Dong, Li and Vuli{\'c}, Ivan and Wei, Furu},
  journal={arXiv preprint arXiv:2501.07542},
  year={2025}
}

@article{xu2025visualplanningletsthink,
  title={Visual Planning: Let's Think Only with Images},
  author={Xu, Yi and Li, Chengzu and Zhou, Han and Wan, Xingchen and Zhang, Caiqi and Korhonen, Anna and Vuli{\'c}, Ivan},
  journal={arXiv preprint arXiv:2505.11409},
  year={2025}
}

@article{chern2025thinking,
  title={Thinking with Generated Images},
  author={Chern, Ethan and Hu, Zhulin and Chern, Steffi and Kou, Siqi and Su, Jiadi and Ma, Yan and Deng, Zhijie and Liu, Pengfei},
  journal={arXiv preprint arXiv:2505.22525},
  year={2025}
}

@inproceedings{goyal2023think,
    title={Think before you speak: Training Language Models With Pause Tokens},
    author={Sachin Goyal and Ziwei Ji and Ankit Singh Rawat and Aditya Krishna Menon and Sanjiv Kumar and Vaishnavh Nagarajan},
    booktitle={The Twelfth International Conference on Learning Representations},
    year={2024}
}

@inproceedings{wang2023guiding,
    title={Guiding Language Model Reasoning with Planning Tokens},
    author={Xinyi Wang and Lucas Caccia and Oleksiy Ostapenko and Xingdi Yuan and William Yang Wang and Alessandro Sordoni},
    booktitle={First Conference on Language Modeling},
    year={2024}
}

@article{cheng2024compressed,
  title={Compressed chain of thought: Efficient reasoning through dense representations},
  author={Cheng, Jeffrey and Van Durme, Benjamin},
  journal={arXiv preprint arXiv:2412.13171},
  year={2024}
}

@article{shen2025codi,
  title={Codi: Compressing chain-of-thought into continuous space via self-distillation},
  author={Shen, Zhenyi and Yan, Hanqi and Zhang, Linhai and Hu, Zhanghao and Du, Yali and He, Yulan},
  journal={arXiv preprint arXiv:2502.21074},
  year={2025}
}

@article{yang2025machine,
  title={Machine Mental Imagery: Empower Multimodal Reasoning with Latent Visual Tokens},
  author={Yang, Zeyuan and Yu, Xueyang and Chen, Delin and Shen, Maohao and Gan, Chuang},
  journal={arXiv preprint arXiv:2506.17218},
  year={2025}
}

@article{li2025latent,
  title={Latent Visual Reasoning},
  author={Li, Bangzheng and Sun, Ximeng and Liu, Jiang and Wang, Ze and Wu, Jialian and Yu, Xiaodong and Chen, Hao and Barsoum, Emad and Chen, Muhao and Liu, Zicheng},
  journal={arXiv preprint arXiv:2509.24251},
  year={2025}
}

@article{pham2025multimodal,
  title={Multimodal Chain of Continuous Thought for Latent-Space Reasoning in Vision-Language Models},
  author={Pham, Tan-Hanh and Ngo, Chris},
  journal={arXiv preprint arXiv:2508.12587},
  year={2025}
}

@article{hurst2024gpt,
  title={Gpt-4o system card},
  author={Hurst, Aaron and Lerer, Adam and Goucher, Adam P and Perelman, Adam and Ramesh, Aditya and Clark, Aidan and Ostrow, AJ and Welihinda, Akila and Hayes, Alan and Radford, Alec and others},
  journal={arXiv preprint arXiv:2410.21276},
  year={2024}
}

@article{zheng2025deepeyes,
  title={DeepEyes: Incentivizing" Thinking with Images" via Reinforcement Learning},
  author={Zheng, Ziwei and Yang, Michael and Hong, Jack and Zhao, Chenxiao and Xu, Guohai and Yang, Le and Shen, Chao and Yu, Xing},
  journal={arXiv preprint arXiv:2505.14362},
  year={2025}
}

@article{wang2024qwen2,
  title={Qwen2-vl: Enhancing vision-language model's perception of the world at any resolution},
  author={Wang, Peng and Bai, Shuai and Tan, Sinan and Wang, Shijie and Fan, Zhihao and Bai, Jinze and Chen, Keqin and Liu, Xuejing and Wang, Jialin and Ge, Wenbin and others},
  journal={arXiv preprint arXiv:2409.12191},
  year={2024}
}

@article{team2024chameleon,
  title={Chameleon: Mixed-modal early-fusion foundation models},
  author={Team, Chameleon},
  journal={arXiv preprint arXiv:2405.09818},
  year={2024}
}

% ============================================================
% Appendix
% ============================================================

\appendix
\section{Training Data Construction}
\label{sec:appendix}

\subsection{Rationale for $N = 4$ Training Stages}
In the {IVT-LR} training, we set the number of stages (N) to 4, which corresponds to three core reasoning steps ($\mathbf{N-1=3}$).  This design choice is not arbitrary; it is motivated by a statistical analysis of the native rationale lengths in the target datasets (M$^3$CoT and ScienceQA).

\begin{figure*}[t]
    \centering
    \includegraphics[width=\linewidth]{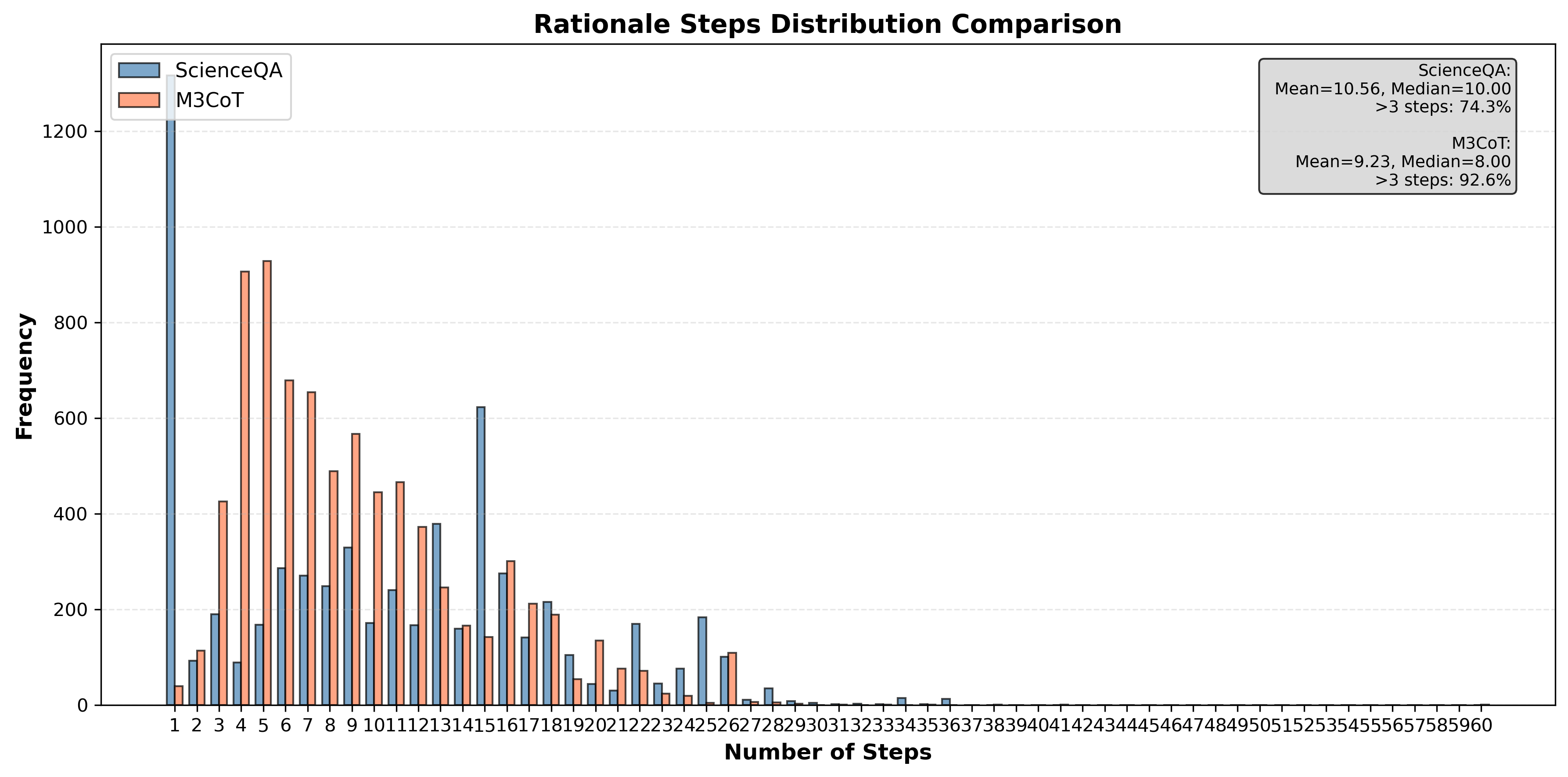}
    \caption{Distribution of native rationale steps across the M3CoT and ScienceQA datasets.}
    \label{fig:rationale}
\end{figure*}

We first examined the distribution of rationale steps (segmented by sentence) in the two datasets. As shown in Figure \ref{fig:rationale}, the median number of rationale steps for both datasets is around 10. Each subtask in the reasoning process usually requires about two to three sentences to complete a causal inference. Thus, a full rationale can be naturally divided into three major reasoning steps, each corresponding to a distinct subtask. Therefore, setting $N=4$ (three reasoning steps) provides a balanced and interpretable abstraction of the overall reasoning process. 

Moreover, the statistical analysis shows that over 70\% of the samples in both datasets contain more than three reasoning steps, and their distributions are highly dispersed. This high dispersion mandates merging adjacent steps for standardization and enhanced computational efficiency. Critically, we simultaneously retain a portion of the original one- and two-step samples. This strategy is essential to preserve the model’s short reasoning ability and bolster generalization across varying reasoning depths, ensuring robust performance regardless of the input's complexity.

\subsection{Examples of Merged Rationales}
Table \ref{tab:question_rationale1} and \ref{tab:question_rationale2} illustrates examples of consolidated rationales with step indices.

\begin{table*}[t]
\centering
\begin{tabularx}{\textwidth}{>{\bfseries}lX}
\toprule
Question & What is the purpose of the hairdryer in the adult's hand? \\
\midrule
Rationale & \textbf{Step 1:} According to the picture, the hair dryer in the adult's hand is not pointed at the hair. This suggests it has other uses besides drying hair. \\
& \textbf{Step 2:} There is a light ball in the air suggests that the air from the hairdryer is holding the ball up. Combined with the dancing little boy in the picture, this shows that the hairdryer is being used to entertain the little boy. \\
& \textbf{Step 3:} Therefore, (C) ``Entertaining the little boy'' is the right answer. \\
\bottomrule
\end{tabularx}
\caption{An example of consolidated rationales for clarity.}
\label{tab:question_rationale1}
\end{table*}

\begin{table*}[t]
\centering
\begin{tabularx}{\textwidth}{>{\bfseries}lX}
\toprule
Question & What is the purpose of the metal pylon on the street near the brick apartment building? \\
\midrule
Rationale & \textbf{Step 1:} The metal pylon on the street indicates that cars are not allowed to drive in the pedestrian area. This inference is derived from the fact that the building next to it is an apartment building which suggests a residential area with high pedestrian traffic. \\
& \textbf{Step 2:} Additionally, the presence of thick white stripes across the street indicates a pedestrian crosswalk. Therefore, it can be concluded that the metal pylon is placed to prevent any intrusion from cars into the area reserved for pedestrians.  \\
& \textbf{Step 3:} Therefore, option B is the correct answer. \\
\bottomrule
\end{tabularx}
\caption{Another example of consolidated rationales for clarity.}
\label{tab:question_rationale2}
\end{table*}

\end{document}